\documentclass[Conference]{IEEEtran}
\usepackage{times}
\usepackage{latexsym}
\usepackage{graphicx}
\usepackage{subfigure}
\usepackage{multirow}
\usepackage{multicol}
\usepackage{stfloats}
\usepackage[justification=centering]{caption}
\usepackage{float}
\usepackage{url}
\usepackage{cite}
\usepackage{booktabs}

\title{WEKA-Based: 
	
	Key Features and Classifier for French of Five Countries}

\author{Zeqian Li, Keyu Qiu, Chenxu Jiao, Wen Zhu, Haoran Tang}

\date{}

\begin{document}
	\maketitle
	\begin{abstract}
		This paper describes a French dialect recognition system that will appropriately distinguish between different regional French dialects. A corpus of five regions - Monaco, French-speaking, Belgium, French-speaking Switzerland, French-speaking Canada and France, which is targeted for construction by the Sketch Engine. The content of the corpus is related to the four themes of eating, drinking, sleeping and living, which are closely linked to popular life. The experimental results were obtained through the processing of a python coded pre-processor and Waikato Environment for Knowledge Analysis (WEKA) data analytic tool which contains many filters and classifiers for machine learning.
	\end{abstract}
	
	\section{Introduction}
	With the development and popularity of artificial intelligence, dialect recognition has become an important text task in the field of data mining\cite{alshutayri-etal-2016-arabic}. There are many French dialects spoken around the World; French dialectologists have studied varieties of local variations, but generally agree these cluster into five main regional dialect: Monaco, Belgium, French-speaking areas in both Switzerland and Canada and France. Since the spatial divide between these countries, the development of French-speaking cultures in these regions has been markedly different. These dialectal differences are worth discovering and studying, not only in terms of cultural knowledge, but also in terms of their importance in the field of computing. For example, dialect recognition can help artificial intelligence to communicate effectively with users from different geographical regions. To find these dialects’ unique features, we use sketch engine to construct dialect corpus and use WEKA's StringToWordVector filter to extract words. What’s more, we used seven classic classifiers --- NaiveBayes, SMO, J48, ZeroR, JRip, c, and NaiveBayesMultinomial to build the model. The use of WEKA's powerful features has led to a series of precise and valuable experimental results. Finally, advantages and performance differences of each classifier are summarized through scientific and quantitative assessment criteria for outcome data.
	
	\section{Business Understanding}
	The objective of this task is to find the characteristics of different regional French dialects to help people identify them accurately. The requirement is to build a corpus of French dialects that meets the requirements by means of the sketch engine. After obtaining the text corresponding to the corpus, to convert it into aircraft rescue and firefighting data file (.arff). Then, using the relevant methods and functions of the WEKA data analytic tool to analyses classified French dialects corpus. Additionally, comparing the performance of different classifiers through the results. The definition of this data mining problem is How to effectively identify the different dialects of French in different regions and help people to be able to accurately determine their differences 
	
	\section{Data Preparation}
	Since Sketch Engine could only save .txt files, which can not be used directly in WEKA. We use Python to clean these data, and process them into .arff format.
	
	The processing procedures and standard are as follows:
	\begin{itemize}
		\item Read the .txt file by lines using Python, and filter out text without \textless p\textgreater  or  \textless /p\textgreater.
		\item Delete the \textless p\textgreater  and  \textless /p\textgreater  labels for each sentence.
		\item Assume that a completed sentence is ended with a full point, exclamation mark or question mark. According to this rule, filter all the unmatched sentence again.
		\item Delete all the sentences with repetition or too few words. 
		\item Use regex to match french characters, remove other symbols.
		\item Add corresponding country label and quotation marks for each sentence.
		\item Add header, composed of attributes, classification, and relationships of data into.arff file applying the format of WEKA data set.
	\end{itemize}
	Thus, a dataset with 10052 instances was constructed. The number of instances was 1674, 1011, 1588, 1478, 1143 for labels Morocco, Belgium,	France, Canada, and Switzerland respectively. Moreover, among these data, the maximum words in a sentence is 1080, while the mean value is 40, and StdDev is 59. As for the test set, The number of instances was 352, 110, 318, 284, 286 for labels with same order as previous. The data statics of training data is shown in the Table 1. 
	
	\begin{table}[htbp]
		\centering
		\setlength{\tabcolsep}{8mm}{
			\begin{tabular}{@{}cccc@{}}
				\toprule
				\textbf{No.} & \textbf{Label} & \textbf{Count} \\ \midrule
				\textbf 1    & {MC}                           & 1674                                              \\
				\textbf 2    & {BE}                           & 1011                                              \\
				\textbf 3    & {FR}                           & 1588                                              \\
				\textbf 4    & {CA}                           & 1478                                              \\
				\textbf 5    & {CH}                           & 1143                                              \\ \bottomrule
		\end{tabular}}
		\caption{\centering Training Data Statistic}
	\end{table}

	\section{Method}
	\subsection{Filter}
	WEKA's StringToWordVector filter was used in this experiment. It has the ability to extract words from the document, and each word was given an attribute of occurrence times.
	\subsection{Modeling}
	Based on WEKA's classifier function and Alrehaili and Atwell's (2017) research on Arabic dialects, seven suitable classifiers, NaiveBayes, SMO, J48, ZeroR, JRip, c, and NaiveBayesMultinomial, were selected for modeling, which effectively avoided blind modeling and thus reduced the experimental time, the result is shown in Table 2.
	
	\begin{itemize}
		\item \textbf{NaiveBayes:} Naive Bayes method is a classification method based on Bayes theorem and the independent assumption of feature conditions, which assumes that each feature of samples is unrelated to other features.
		\item \textbf{SMO:} SMO classifier implements the sequential minimum optimization algorithm, which can be used to train support vector classifiers. This implementation replaces all missing values globally and converts nominal attributes to binary attributes \cite{6789464}.
		\item \textbf{J48:} J48 classifier implements the	C4.5 algorithm for building decision trees which were pruned or unpruned. A pruned C4.5 decision tree was chosen in this experiment.
		\item \textbf{ZeroR:} ZeroR classifier relies on the target and ignores all predictions. It only predicts most classes and has no predictability.
		\item \textbf{JRip} JRip classifier implements a propositional rule learner, Repeated Incremental Pruning to Produce Error Reduction (RIPPER) \cite{article}.
		\item \textbf{REPTree:} REPTree classifier is a fast decision tree learner. It trims the decision tree using error reduction pruning.
		\item \textbf{NaiveBayesMultinomial:} NaiveBayesMultinomial is a classifier for building and using a multinomial Naive Bayes classifier.
	\end{itemize}
	
	\begin{table*}[]
		\center
		\setlength{\tabcolsep}{16mm}{
			\begin{tabular*}{\textwidth}{@{}cccc@{}}
				\toprule
				\textbf{Classifier} & \textbf{\begin{tabular}[c]{@{}c@{}}Evaluate on \\ Training Set\end{tabular}} & \textbf{\begin{tabular}[c]{@{}c@{}}10 Fold Cross \\ Validation\end{tabular}} & \textbf{\begin{tabular}[c]{@{}c@{}}60\% Train\\ 40\% Test\end{tabular}} \\ \midrule
				\textbf{NaiveBayes}            & 37.0683                           & 35.6849                           & 33.9229                        \\
				\textbf{SMO}                   & 89.8607                           & 59.2882                           & 57.0222                        \\
				\textbf{J48}                   & 76.8297                           & 49.6990                           & 45.0106                        \\
				\textbf{ZeroR}                 & 25.1875                           & 25.1875                           & 25.0264                        \\
				\textbf{JRip}                  & 37.2162                           & 34.2592                           & 33.8173                        \\
				\textbf{REPTree}               & 58.6651                           & 48.1571                           & 44.7994                        \\
				\textbf{NaiveBayesMultinomial} & 65.1178                           & 59.0981                           & 57.8669                        \\ \bottomrule
		\end{tabular*}}
		\caption{\centering The accuracy of different classifiers}
	\end{table*}
	
	\subsection{Evaluating the performance of models}
	In this experiment, three validation method provided by WEKA were used to evaluate the performance of the eight classifiers: use the training dataset, 10-fold cross-validation, and percentage split which divides the training set into 60\% for training and 40\% for testing.
	\begin{itemize}
		\item \textbf{Evaluating on training set:} A method that the training set and the test set use the same data.
		\item \textbf{Cross-Validation:} Divide the training set into N parts, use n-1 part for training, use 1 part for testing, and so on for N times, and finally calculate the overall result. N was set to 10 in this experiment.
		\item \textbf{Percentage Split:} Divide the training set into two parts in proportion, one for training and one for testing. In this experiment, the training set was divided into 60\% and 40\% where 60\% for training and 40\% for testing.
	\end{itemize}
	
	\subsection{Feature}
	The first experiment which is to choose the best classifier to recognize French dialects show that SMO is the best machine learning classification algorithm, but we could still improve the accuracy by adjusting the parameters and features. The WordTokenizer setting assumes that the feature is strings or words between spaces while CharacterNGramTokenizer assumes features are 1, 2 or 3 character sequences. NGramTokenizer tokenize the inputs into different kinds of n-grams. The input can be a character vector of any length. We tried to set Max and Min both to 1 gives a model based on single characters; Max and Min both to 2 gives us a bigram model; Max and Min both to 3 is a trigram model. Moreover, we also tried to use WordTokenizer with TF and IDF both False and NGramTokenizer with Max to 3 and Min to 1. 
	
	The result of different tokenizer is shown in the Table 3.
	
	\begin{table*}[]
		\centering
		\setlength{\tabcolsep}{23mm}{
			\begin{tabular*}{\textwidth}{@{}ccc@{}}
				\toprule
				\textbf{Features}          & \textbf{Evaluate on Training Set} & \textbf{60\% Train, 40\% Test} \\ \midrule
				\textbf{WordTokenizer}     & 89.8607                           & 57.0222                       \\
				\textbf{NGramTokenizer}    & 89.9043                           & 56.5693                       \\
				\textbf{Character Unigram} & 34.9289                           & 33.6113                       \\
				\textbf{Character Bigram}  & 70.5250                           & 41.3659                       \\
				\textbf{Character Trigram} & 94.1979                           & 45.6490                       \\ \bottomrule
		\end{tabular*}}
		\caption{\centering The accuracy of SMO classifier with different features}
	\end{table*}

	\section{Result}
	Finally, we used the prepared separate test data set to evaluate our system via SMO classifier and Character NGramTokenizer filter. Our experimental result is obtained by using TriGramTokenizer, IDF=False, TF=False. The total number of the instances is 1350, the accuracy of our model approximately equals to 39\%, and the Mean Absolute Error (MAE) indicating the average value of the distance between the predicted value of the model and the real value of the sample is 29\%.
	
	\begin{table}[htbp]
		\centering
		\setlength{\tabcolsep}{4.8mm}{
			\begin{tabular}{|c|cc|}
				\hline
				\textbf{Result}                    & \multicolumn{1}{l|}{\textbf{Amount}} & {\textbf{Rate}} \\ \hline
				\textbf{Total Number of Instances} & \multicolumn{1}{c|}{1350}            & /                                  \\ \hline
				\textbf{Correctly Classified}      & \multicolumn{1}{c|}{531}             & 39.35\%                          \\ \hline
				\textbf{Incorrectly Classified}    & \multicolumn{1}{c|}{819}             & 60.65\%                          \\ \hline
				\textbf{Mean Absolute Error}       & \multicolumn{2}{c|}{0.2908}                                               \\ \hline
			\end{tabular}
			\caption{SMO Classified Result}}
	\end{table}
	
	\begin{itemize}
		\item \textbf{F1-Score:} Precision (PRE) and Recall (REC) are two significant and common metrics when evaluating the effectiveness of binary classification model. However, under certain circumstances, Precision and Recall indicators sometimes affect each other. F1-Measure is generally a harmonic method balancing and combining both Precision and Recall metrics, which is calculated as followed:
		
		\begin{table*}[H]
			\centering
			\setlength{\tabcolsep}{3mm}{
				\begin{tabular}{ccccc}
					\hline
					\textbf{Class}        & \textbf{Precision} & \textbf{Recall} & \multicolumn{1}{l}{\textbf{F1-Score}} & \multicolumn{1}{l}{\textbf{AUC}} \\ \hline
					\textbf{MC}           & 0.426              & 0.625           & 0.506                                 & 0.716                            \\
					\textbf{BE}           & 0.179              & 0.099           & 0.127                                 & 0.643                            \\
					\textbf{FR}           & 0.386              & 0.446           & 0.414                                 & 0.612                            \\
					\textbf{CA}           & 0.440              & 0.345           & 0.386                                 & 0.706                            \\
					\textbf{CH}           & 0.348              & 0.184           & 0.241                                 & 0.676                            \\
					\textbf{Weighted Avg} & 0.377              & 0.393           & 0.373                                 & 0.673                            \\ \hline
			\end{tabular}}
			\caption{\centering The values of various indexes of SMO classifier}
		\end{table*}
		
		$$\frac{1}{F_\beta} = \frac{1}{1+\beta^{2}}(\frac{1}{P} + \frac{\beta^{2}}{P})$$  $$F_{\beta} = \frac{(\beta^{2}+1)\times P\times R}{(\beta^{2}\times P) + R }$$ Where $\beta$ is the parameter, P is Precision and R is Recall. Normally, we use F1-Score method, which is derived from setting $\beta$ to 1. The range of F1-Score is from 0 to 1, with 0 represents the worst output. The performance of the model gets better with the increase of F1-Score.
		\item \textbf{ROC Curve:} The abscissa of the ROC plane is false positive rate (FPR), and the ordinate is true positive rate (TPR). For a certain classifier, we can obtain a TPR and FPR dotted pair according to its performance on the test sample. In this way, the classifier can be mapped to a point on the ROC plane. By adjusting the threshold used by this classifier, we can get a curve passing through (0, 0), (1, 1), which shapes the ROC curve of this classifier.
		
		\item \textbf{AUC: } AUC is a standard metric used to measure the quality of classification model. The value of AUC is the size of the area below the ROC curve. Generally, the value of AUC is between 0.5 and 1.0, and a larger AUC value represents the better performance of the model\cite{article-one}.
	\end{itemize}
	Table 5 shows the respective and average accuracy of Precision, Recall, F1-Score and AUC of five identical classes (i.e.MC, BE, FR, CA, CH) using SMO classifier.

	\section{Conclusion}
	Our modified system was built with WEKA data analytic tool and SMO classifier after measuring and comparing the performance of different classifiers tested on training set (around 94\% accuracy) and 60\% for training while the remaining 40\% for testing (about 57\% accuracy at the best situation). After selecting SMO as the primary chosen classifier, we then evaluated our system via SMO classifier and various filter using separate data set. We have figured out that with our dataset, Character TriGramTokenizer have the best performance. We found that the reason that we get a lower accuracy is the low distinction of text and the unselected data, since we used the corpus from Sketch Engine. Obviously, if the training data set contains mislabeled or appropriated data, the accuracy of classifier could achieve will be reduced.
	
	\bibliographystyle{ieeetr}
	\bibliography{document}

\begin{thebibliography}{1}

\bibitem{alshutayri-etal-2016-arabic}
A.~Alshutayri, E.~Atwell, A.~Alosaimy, J.~Dickins, M.~Ingleby, and J.~Watson,
  ``{A}rabic language {WEKA}-based dialect classifier for {A}rabic automatic
  speech recognition transcripts,'' in {\em Proceedings of the Third Workshop
  on {NLP} for Similar Languages, Varieties and Dialects ({V}ar{D}ial3)},
  (Osaka, Japan), pp.~204--211, The COLING 2016 Organizing Committee, Dec.
  2016.

\bibitem{6789464}
S.~S. Keerthi, S.~K. Shevade, C.~Bhattacharyya, and K.~R.~K. Murthy,
  ``Improvements to platt's smo algorithm for svm classifier design,'' {\em
  Neural Computation}, vol.~13, no.~3, pp.~637--649, 2001.

\bibitem{article}
W.~Cohen, ``Fast effective rule induction,'' {\em Twelfth International
  Conference on Machine Learning: 1995}, vol.~95, 10 2000.

\bibitem{article-one}
E.~Atwell, C.~Brierley, C.~Rowland, and J.~Anderson, ``Semantic pathways: a
  novel visualisation of varieties of english,'' {\em . ICAME Journal of the
  International Computer Archive of Modern English}, vol.~37, pp.~5--36, 01
  2013.

\end{thebibliography}

\end{document}